\documentclass[11pt,letterpaper]{article}
\usepackage{liveaigc_report}
\ifPDFTeX
  \microtypesetup{tracking=false,kerning=false,spacing=false}
\fi
\usepackage{mathtools,xspace}
\definecolor{rankone}{RGB}{235,101,101}
\definecolor{ranktwo}{RGB}{247,161,161}
\definecolor{rankthree}{RGB}{253,218,218}
\title{\fontsize{20pt}{24pt}\selectfont Taomate-F1: Trajectory-Aware Streaming\\Audio-Visual Generation from First and Last Frames}
\reportauthors{Chenwei Zhang\textsuperscript{1,*}, Qijun Gan\textsuperscript{1,*}, Xinyang Li\textsuperscript{1},
Zichao Yu\textsuperscript{2}, Meiguang Jin\textsuperscript{1,\textdagger}, Junfeng Ma\textsuperscript{1}}
\authornote{\textsuperscript{*}Equal contribution.\quad \textsuperscript{\textdagger}Corresponding author.}
\affiliation{\textsuperscript{1}Alibaba Group\qquad \textsuperscript{2}The University of Hong Kong}
\teamname{Project page\quad \url{https://gatswei.github.io/Taomate-F1-Demo/}}
\shorttitle{Taomate-F1}

\reportlabel{Technical Report}
\reportdate{November 19, 2026}
\leftlogo[height=0.9cm]{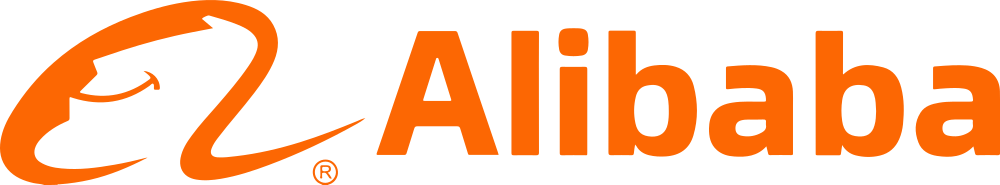}
\rightlogo[height=0.9cm,trim=436bp 638bp 306bp 480bp,clip]{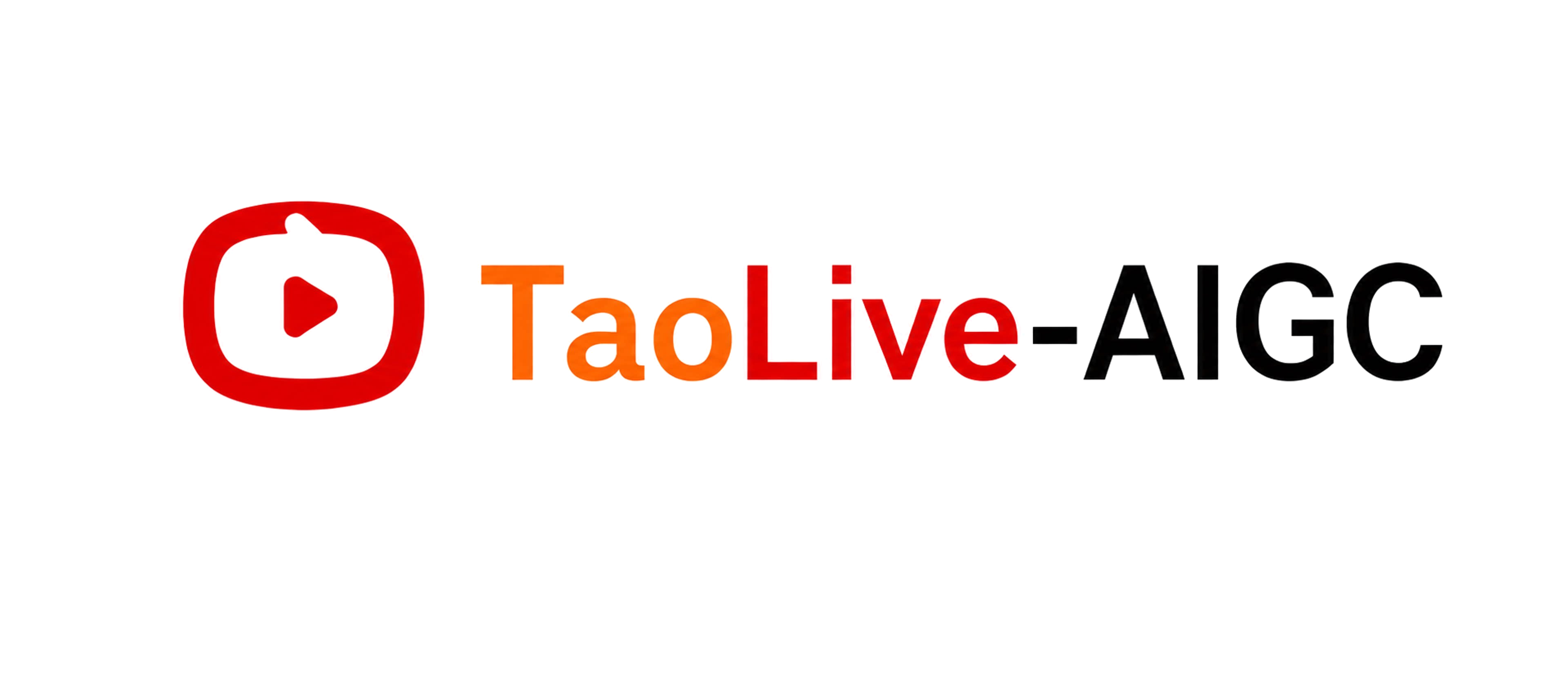}

\hypersetup{
  pdftitle={Taomate-F1: Trajectory-Aware Streaming Audio-Visual Generation from First and Last Frames},
  pdfauthor={Chenwei Zhang, Qijun Gan, Xinyang Li, Zichao Yu, Meiguang Jin, Junfeng Ma}
}

\begin{document}
\maketitle
\begin{abstract}
Interactive digital humans and live storytelling require real-time audiovisual streams whose visual evolution remains controllable over long durations. First and last frames guide each segment toward a specified visual state, while linking successive transitions through shared boundary states supports smooth long-video continuation. The challenge is to reach these targets while preserving temporal continuity, since emitted content becomes irreversible history that constrains subsequent generation. We introduce \emph{Taomate-F1}, which, to the best of our knowledge, is the first method for streaming first-and-last-frame-to-video-and-audio (FL2VA) generation. Given two boundary images, text, and a duration, our model causally emits synchronized audiovisual blocks. During distillation, we encode the target image at its global terminal time using rotary position embeddings and combine endpoint reconstruction with supervision of intermediate spatial structure and temporal change from paired teacher trajectories. Full-chain gradients propagate this supervision through every retained denoising step within each block. We then refine the distilled generator through reinforcement learning with Cross-Block Group Relative Policy Optimization (CB-GRPO). Building on consistency-model policy optimization, CB-GRPO propagates reward-weighted losses from later blocks through selected history dependencies into earlier content generation. Quality and terminal-image rewards define preferences over complete rollouts. On 168 one-minute transitions, Taomate-F1 leads on eight of nine quality metrics, including temporal stability and color consistency. At ten seconds, its quality is comparable to the bidirectional base model.

\end{abstract}
\begin{figure}[!htbp]
\centering\includegraphics[width=.97\textwidth,trim=0 0 0 18.3bp,clip]{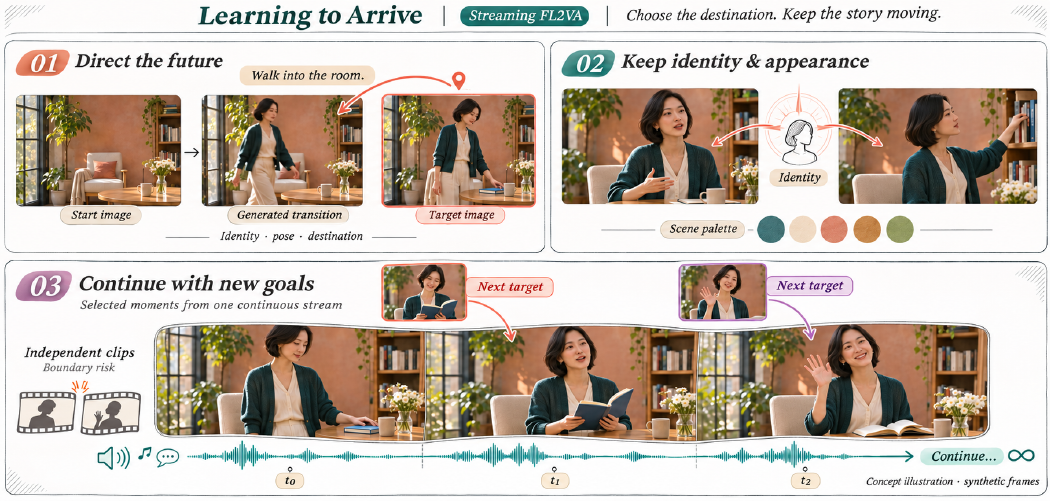}
\captionsetup{hypcap=false}
\caption{\textbf{Streaming FL2VA for controllable, open-ended audiovisual generation.}
First and last images provide concrete visual references for directing story development (top left) and anchoring identity, color, and other appearance attributes (top right). Successive targets enable continuous generation with controllable intermediate keyframes (bottom), avoiding motion discontinuities at joins and cumulative color drift.}
\label{fig:teaser}
\end{figure}

\section{Introduction}

Real-time streaming audio-video generation supports interactive digital humans and live storytelling by producing synchronized sound and motion in response to evolving input. TaoMate~\cite{gan2026taomate} and OmniForcing~\cite{omniforcing2026} emit audiovisual blocks conditioned on generated history, allowing playback before generation finishes. Their bounded active context permits continuation without a fixed session length, motivating interfaces that let users direct the evolving stream.

Streaming first-and-last-frame-to-video-and-audio (FL2VA) generation offers such an interface. Given first and last images, text, and a duration, it connects specified visual states through causal audiovisual blocks. Streaming text-to-video-and-audio (T2VA) guides semantic content through prompts but cannot directly impose the visual state at a future time. The first image establishes the current context, while the last image specifies a concrete destination. Together, they provide references for scene layout, character appearance, object arrangement, pose, and other attributes that text alone leaves ambiguous. These paired visual references narrow plausible continuations and guide story development toward the requested outcome at the specified time (Figure~\ref{fig:teaser}, top left).

Streaming T2VA also lacks direct visual constraints on identity and scene colors. Our method anchors these attributes to the requested images throughout the transition, suppressing identity and color drift (Figure~\ref{fig:teaser}, top right). Successive targets become controllable intermediate keyframes, with each new target directing the next part of the same stream (Figure~\ref{fig:teaser}, bottom). Bidirectional FL2VA methods can connect prescribed images within individual clips, but matching a boundary pose does not enforce matching motion speed. A slow action can therefore accelerate abruptly after a join. Reusing generated boundary frames also carries uncorrected appearance errors into subsequent clips, where deviations can accumulate into color drift. With streaming FL2VA, users can specify these intermediate visual states while the sequence continues.

A causal stream must reach its destination through a history it cannot revise. Whole-clip audiovisual models such as MiniMax-H3 (H3 Base)~\cite{minimax2026h3modelcard} and H3 Turbo~\cite{lightx2v2026h3turbo} jointly process a prescribed temporal span. A stream instead commits each block as both visible content and future context. Early deviations change the route to the target as the remaining transition time shrinks. Even a locally plausible continuation may lead away from the destination. Training must therefore connect intermediate generation decisions to eventual arrival while preserving a coherent trajectory.

Taomate-F1 generates synchronized audiovisual blocks toward successive visual targets. Users can specify each new destination during playback. Three training choices connect the visual target to the blocks that precede it.

\emph{Temporal Conditioning and Trajectory Supervision} places the target image at a fixed global terminal coordinate using rotary position embedding (RoPE)~\cite{su2024roformer}. Advancing query coordinates encode the time remaining to this target. For each student rollout paired with a frozen teacher under shared conditions and initial noise, we match pooled video latents and temporal differences, alongside endpoint reconstruction. These video losses complement distribution matching distillation (DMD~\cite{yin2024dmd} and DMD2~\cite{yin2024dmd2}), while audio retains the base distillation supervision.

\emph{Full-Chain Denoising Gradients} propagate these losses through every retained denoising step within each block, training early computations through their effect on the output. Following Self Gradient Forcing (SGF)~\cite{zhuang2026selfgradientforcing}, later losses also update historical key--value writing while generated history latents remain fixed on this path.

\emph{Cross-Block Policy Learning} uses our Cross-Block Group Relative Policy Optimization (CB-GRPO) to propagate reward-weighted losses from later blocks through selected history dependencies into earlier content generation. Each block is thus trained both as present content and as context for the continuing stream.

\begin{itemize}
    \item \textbf{The first streaming FL2VA method.}
    Causal joint audio--video generation under first- and last-frame conditions enables visual destination control during generation and playback.

    \item \textbf{Temporal conditioning and trajectory supervision.}
    We combine global target-time conditioning with paired trajectory and endpoint supervision, and train every retained denoising step through full-chain gradients.

    \item \textbf{Cross-Block Policy Learning with CB-GRPO.}
    Later-block policy feedback trains earlier generated content for both its own output and its effect on the continuing stream.

    \item \textbf{Long-horizon quality and inter-frame consistency.}
    Experiments demonstrate better preservation of visual quality over long sequences and stronger inter-frame consistency.
\end{itemize}

\section{Related work}

\textbf{Video and audiovisual generation.}
DDPM~\cite{ho2020ddpm}, DDIM~\cite{song2020ddim}, latent diffusion~\cite{rombach2021ldm}, and DiT~\cite{peebles2022dit} provide foundations for video diffusion. Video Diffusion Models~\cite{ho2022videodiffusion} extend denoising to temporal data, and Make-A-Video~\cite{singer2022makeavideo} transfers text--image priors to video through motion learning. VideoLDM~\cite{blattmann2023videoldm} and AnimateDiff~\cite{guo2023animatediff} add temporal modeling to image backbones, while CogVideoX~\cite{yang2024cogvideox} and Open-Sora 2.0~\cite{zheng2025opensora2} train video transformers.

Image conditioning specifies visual states along a trajectory. DynamiCrafter~\cite{xing2023dynamicrafter} animates a single reference image, while PixelDance~\cite{zeng2023pixeldance} and Wan2.1~\cite{wan2025technical,wan2025repository} use both first and last frames. SEINE~\cite{chen2023seine} uses masked video diffusion for transition modeling and prediction, while SparseCtrl~\cite{guo2023sparsectrl} introduces sparse temporal controls. ToonCrafter~\cite{xing2024tooncrafter} and Generative Inbetweening~\cite{wang2024generativeinbetweening} use video priors to interpolate between reference images.

MM-Diffusion~\cite{ruan2022mmdiffusion} and AV-DiT~\cite{wang2024avdit} jointly generate audio and video, and Movie Gen~\cite{polyak2024moviegen} develops generators for both modalities. EMO~\cite{tian2024emo}, VASA-1~\cite{xu2024vasa}, and Hallo~\cite{xu2024hallo} animate portraits from supplied audio. MiniMax H3 Base~\cite{minimax2026h3modelcard} and its few-step Turbo variant~\cite{lightx2v2026h3turbo} generate complete audiovisual clips from first and last frames. We use H3's joint audiovisual representation to generate successive blocks toward a prescribed visual destination.

\textbf{Autoregressive generation and forcing.}
Autoregressive models extend a committed prefix. VideoPoet~\cite{kondratyuk2023videopoet} applies decoder-only modeling to multimodal generation. Diffusion Forcing~\cite{chen2024diffusionforcing} couples causal sequence modeling with token-wise noising. SkyReels-V2-DF~\cite{chen2025skyreelsv2} uses Diffusion Forcing to extend video beyond a single clip. Progressive distillation~\cite{salimans2022progressive}, consistency models~\cite{song2023consistency}, latent consistency models~\cite{luo2023lcm}, and distribution matching distillation (DMD~\cite{yin2024dmd} and DMD2~\cite{yin2024dmd2}) support few-step sampling. CausVid~\cite{yin2024causvid} distills a bidirectional video teacher into a causal student using distribution matching.

Self Forcing~\cite{huang2025selfforcing} trains on student rollouts and truncates gradients to limit memory use. LongLive~\cite{yang2025longlive} and Self-Forcing++~\cite{cui2025selfforcingplusplus} extend rollout-based training to longer sequences through segment-level supervision. Extending self-forcing to joint audiovisual generation, OmniForcing~\cite{omniforcing2026} distills a bidirectional teacher into a streaming generator trained on its own audiovisual history. TaoMate~\cite{gan2026taomate} retains fixed-capacity audiovisual memory beyond its active context, combining a fixed initial reference with evolving content states. Self Gradient Forcing~\cite{zhuang2026selfgradientforcing} uses future losses to train memory writing while stopping gradients through generated content. Self-Replay Forcing in Vidu S2~\cite{zhang2026vidus2} re-noises and differentiably replays segments of detached student rollouts. Connected Self Forcing~\cite{zhang2026connectedselfforcing} replays historical key--value writing and earlier generation. We extend gradient propagation to the full denoising chain within each audiovisual block, allowing the loss on the completed block to train earlier denoising steps. Taomate-F1 is the first training framework for autoregressive FL2VA.

\textbf{Reward-based generative-model refinement.}
Proximal policy optimization~\cite{schulman2017ppo} uses a surrogate objective, while group-relative policy optimization~\cite{shao2024deepseekmath} estimates advantages across samples under the same condition. DDPO~\cite{black2023ddpo} and DPOK~\cite{fan2023dpok} apply policy gradients to denoising. AlignProp~\cite{prabhudesai2023alignprop} and DRaFT~\cite{clark2024draft} backpropagate differentiable rewards through denoising, and Diffusion-DPO~\cite{wallace2023diffusiondpo} learns from pairwise preferences. Flow-GRPO~\cite{liu2025flowgrpo} uses stochastic flow transitions, while DiffusionNFT~\cite{zheng2025diffusionnft} uses a reward-weighted forward-process objective. RAVEN~\cite{lu2026raven} presents a framework for causal video extrapolation and proposes CM-GRPO for online reinforcement learning. CM-GRPO uses a stopped target to realize reward-weighted score gradients for Gaussian consistency transitions. Our CB-GRPO further adapts CM-GRPO~\cite{lu2026raven} for streaming FL2VA. We backpropagate policy losses from later blocks through selected paths in the generated history to train earlier content generation according to its influence on subsequent blocks.

\FloatBarrier
\section{Method}
\label{sec:method}

\begin{figure}[!htbp]
\centering
\includegraphics[width=.96\textwidth]{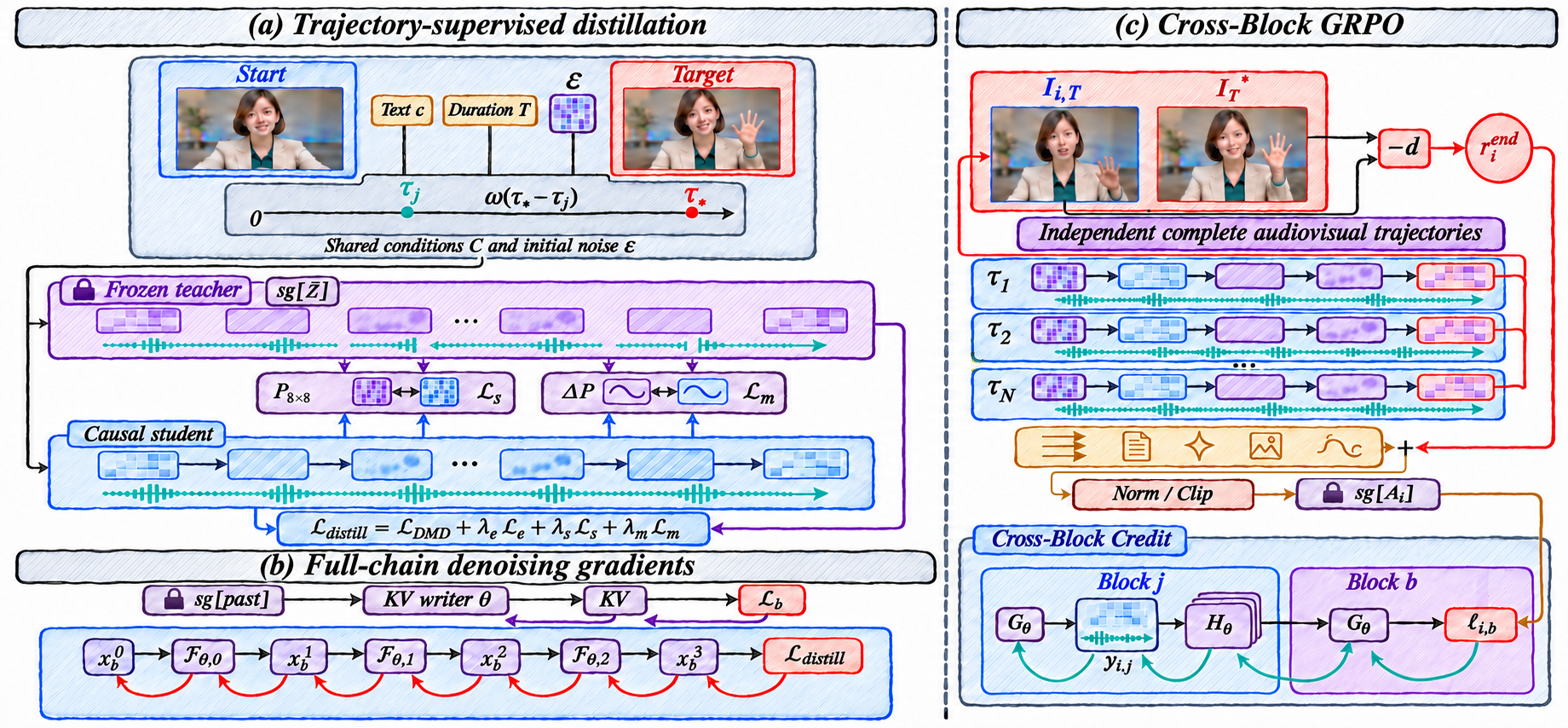}
\caption{\textbf{Taomate-F1.} \textbf{(a) Temporal conditioning and trajectory supervision.} Target-image RoPE encodes the global terminal time. Paired teacher trajectories supervise intermediate structure and temporal change alongside endpoint reconstruction and DMD. \textbf{(b) Full-chain denoising gradients.} Output losses train every retained denoising step within each block. \textbf{(c) Cross-block policy learning.} CB-GRPO propagates reward-weighted later-block losses through selected history paths into earlier content generation. Quality and target-image rewards define rollout preferences.}
\label{fig:overview}
\end{figure}

Taomate-F1 trains a causal audiovisual generator to connect prescribed first and last images over a requested duration. Unlike jointly denoised clips, a stream cannot revise emitted blocks. Any deviation in pose, layout, or motion becomes history for subsequent generation, which must reach the same destination in less remaining time.

Relative to text-to-video-and-audio (T2VA) generation, FL2VA adds destination consistency to error accumulation. A locally plausible trajectory may satisfy a prompt yet no longer connect smoothly to the target image. Continuing it can miss the destination, while correcting it late can force an abrupt transition. Long-horizon training must therefore preserve both appearance and compatibility with the destination.

Figure~\ref{fig:overview} shows how the target guides the trajectory during distillation and how later policy losses train earlier blocks.

\subsection{Temporal conditioning and trajectory supervision}
\label{sec:trajectory}

The requested images and duration specify the visual endpoints and terminal time of a transition, but leave the intervening trajectory underdetermined. Temporal conditioning preserves H3's native target-frame timing across causal blocks, associating the requested visual state with the transition's global endpoint. We complement this temporal conditioning with direct supervision from a paired teacher rollout, providing explicit targets for intermediate spatial structure and temporal change as causal generation proceeds.

For one requested transition, let \(C=(I_0^\star,I_T^\star,c,T)\) contain the first image, target final image, text condition, and global terminal frame associated with the requested duration. Using the joint video--audio latent representation of H3~\cite{minimax2026h3modelcard}, the generator emits audiovisual blocks \(Y_b=(V_b,A_b)\), \(b=1,\ldots,B\). Each block reads clean history \(h_b\) from the committed prefix. The terminal index \(T\) is defined over the complete transition and does not reset at block boundaries.

\noindent\textbf{Temporal Conditioning.}
Following H3's native keyframe conditioning~\cite{minimax2026h3modelcard}, we place reference-image tokens on its audiovisual timeline using rotary position embedding (RoPE)~\cite{su2024roformer}. The first image is anchored at the media time origin. Target-image patches retain their spatial coordinates and share \(\tau_\star\), determined by the video latent length of the complete transition. We retain this endpoint coordinate across all causal blocks and internal conditioning windows, while generated queries advance along the transition's timeline. For a query at \(\tau_j\), the temporal query--target rotation at frequency \(\omega\) is \(\omega(\tau_\star-\tau_j)\). This preserves the relative temporal displacement to the requested target throughout generation, while trajectory supervision trains the intermediate progression toward it.

\noindent\textbf{Intermediate Trajectory Supervision.}
If the subject turns away from the target pose or drifts out of frame midway, subsequent blocks inherit that change. As the remaining interval shrinks, the generator can miss the requested endpoint or reach it through an abrupt change in pose or framing. To guide the approach to the target, we pair the student's video latents \(Z\), assembled over the transition, with a trajectory \(\bar Z\) from a frozen teacher under the same condition \(C\) and initial noise. Following the shared-noise teacher regression used in distribution matching distillation (DMD)~\cite{yin2024dmd}, we use the teacher trajectory directly as a regression target and stop gradients through the teacher output.

To supervise intermediate structure, spatial average pooling \(P\) maps each video latent to an \(8\times8\) grid while preserving its channels and temporal order. We define \((\Delta P(Z))_k=P(Z)_{k+1}-P(Z)_k\), where \(k\) indexes consecutive video latent times across the assembled transition. The difference is taken at corresponding channels and spatial locations. The structure and temporal-change losses are
\begin{equation}
\begin{aligned}
\mathcal L_s
&=
\operatorname{SmoothL1}\!\left(
P(Z)_{\mathrm{inter}},
P(\operatorname{sg}[\bar Z])_{\mathrm{inter}}
\right),\\
\mathcal L_m
&=
\operatorname{SmoothL1}\!\left(
\Delta P(Z),
\Delta P(\operatorname{sg}[\bar Z])
\right),
\end{aligned}
\label{eq:trajectory_losses}
\end{equation}
where \(\operatorname{sg}\) denotes stop-gradient. The subscript \(\mathrm{inter}\) excludes the first and last two video latent times from \(\mathcal L_s\), while \(\mathcal L_m\) uses all adjacent time pairs. Both losses average elementwise SmoothL1 errors over the compared tensor entries, with transition parameter \(\beta=1\). Thus \(\mathcal L_s\) matches intermediate spatial structure, and \(\mathcal L_m\) matches its temporal evolution.

At the two endpoints, we supervise the decoded student frames directly against the requested images,
\begin{equation}
\mathcal L_e
=
\frac{1}{2}
\sum_{e\in\{0,T\}}
\operatorname{mean}
\operatorname{SmoothL1}\!\left(
S(I_e)-S(I_e^\star)
\right),
\label{eq:method1}
\end{equation}
where \(I_e\) is the decoded student image at endpoint \(e\), and \(I_e^\star\) is the corresponding requested image. The generated and reference RGB images are geometry-matched and represented in \([0,1]\). The operator \(S\) performs antialiased bilinear resizing to \(128\times128\). Endpoint extraction follows the decoder's native temporal geometry, and the loss is evaluated on the generated frames.

The complete distillation objective is
\begin{equation}
\mathcal L_{\mathrm{distill}}
=
\mathcal L_{\mathrm{DMD}}
+
\lambda_e\mathcal L_e
+
\lambda_s\mathcal L_s
+
\lambda_m\mathcal L_m.
\label{eq:method3}
\end{equation}
The auxiliary losses supervise video. Audio receives the base distillation loss.

\subsection{Full-chain denoising gradients}
\label{sec:credit}

The distillation objective in Eq.~\ref{eq:method3} evaluates completed generated outputs, whose values depend on the composition of multiple denoising steps. To let this supervision account for the contribution of each step, we differentiate the complete retained denoising chain within each block. Endpoint and trajectory errors can then train early denoising computations through their influence on the emitted content. DiffusionCLIP~\cite{kim2022diffusionclip} and Direct Reward Fine-Tuning (DRaFT)~\cite{clark2024draft} explore differentiation through iterative sampling. We apply this principle to the few-step sampler used for streaming distillation.

For block \(b\), let \(x_b^k\) denote its joint video--audio latent state after \(k\) of the \(K\) retained denoising transitions. With initial state \(x_b^0\), shared parameters \(\theta\), and history \(h_b\), the generated block is obtained by
\begin{equation}
x_b^{k+1}=F_{\theta,k}(x_b^k,h_b,C),\qquad y_b=x_b^K.
\label{eq:denoising_composition}
\end{equation}
Here \(k=0,\ldots,K-1\). Let \(q_{b,K}\) denote the direct output gradient supplied to \(y_b\) by Eq.~\ref{eq:method3}. Define the local Jacobians
\(J^x_{b,k}=\partial F_{\theta,k}/\partial x_b^k\)
and
\(J^\theta_{b,k}=\partial F_{\theta,k}/\partial\theta\).
For this within-block contribution, the history \(h_b\) is held fixed. Backpropagation through the composition gives
\begin{equation}
\begin{aligned}
q_{b,k} &= (J^x_{b,k})^\top q_{b,k+1},\\
g_b^{\mathrm{denoise}} &= \sum_{k=0}^{K-1}(J^\theta_{b,k})^\top q_{b,k+1}.
\end{aligned}
\label{eq:denoising_gradient}
\end{equation}

The recurrence runs from \(k=K-1\) to \(0\), accumulating a gradient for the shared parameters at each step. Self Forcing~\cite{huang2025selfforcing} detaches the states preceding a sampled exit. Our final output loss reaches all \(K\) steps that produced the block.

During distillation, each block reads student-generated history, as it does at inference. Following SGF~\cite{zhuang2026selfgradientforcing}, later losses also train the key--value writer that encodes earlier blocks. On this path, the generated historical latents are stop-gradient inputs. The writer gradient and the within-block gradient in Eq.~\ref{eq:denoising_gradient} both update the shared parameters.

\subsection{Cross-block policy learning}
\label{sec:policy}

An emitted block in a causal stream is both visible content and context for future generation. Its effect on the completed transition therefore extends beyond its own prediction. An early pose or appearance change alters the history from which later blocks must reach the requested endpoint. We introduce Cross-Block GRPO (CB-GRPO) to train this dependency. Later policy losses propagate through selected history paths into the computations that generated earlier content, allowing a block to be optimized for its influence on subsequent decisions.

Starting from the distilled generator, we sample independent complete audiovisual rollouts for each condition \(C\). Candidate \(i\) generates its own prefix, which supplies history \(h_{i,b}\) to block \(b\). A stopped group-relative advantage \(A_i\) expresses the quality and terminal-image agreement of the complete rollout. The same advantage weights each local policy loss along that rollout.

CB-GRPO uses CM-GRPO~\cite{lu2026raven}, the consistency-model policy objective introduced by RAVEN, to define the local loss. For a recorded denoising transition with noise levels \(u>s>0\), let \(\hat x_\theta\) be the predicted clean latent and \(p_\theta(z_s\mid z_u,h_{i,b},C)\) the conditional Gaussian kernel of the consistency sampler. The stopped-target loss \(\ell_{i,b}\) produces the update
\begin{equation}
\nabla_{\hat x_\theta}\ell_{i,b}
=
-A_i\nabla_{\hat x_\theta}
\log p_\theta(z_s\mid z_u,h_{i,b},C).
\label{eq:policy_local_score}
\end{equation}
We apply this objective to both video and audio at a sampled nonterminal denoising time.

\noindent\textbf{Cross-block feedback.}
We extend the policy update through the generated content that later blocks consume. RAVEN~\cite{lu2026raven} allows later losses to update the representations of sampled clean history. We retain this representation-level path and further differentiate selected computations that produced the historical content. Let \(y_{i,j}\) be the clean audiovisual output of an earlier block \(j<b\). Its contribution to the update from the later loss \(\ell_{i,b}\) is
\begin{equation}
g_{i,j\leftarrow b}
=
\left(
\frac{\partial y_{i,j}}{\partial\theta}
\right)^\top
\left(
\frac{\partial h_{i,b}}{\partial y_{i,j}}
\right)^\top
\frac{\partial\ell_{i,b}}{\partial h_{i,b}}.
\label{eq:method13}
\end{equation}
The history derivative carries the later loss back to the earlier output \(y_{i,j}\). Its generator Jacobian then updates the parameters that produced that output. We add this contribution to the local policy and history-encoder gradients.

Following stochastic computation graphs~\cite{schulman2015scg}, we hold recorded stochastic states, rewards, and advantages fixed and differentiate the selected deterministic paths. A zero-noise transition contributes through its clean output when a later block uses it as history. The local Gaussian score loss uses positive noise levels.

\noindent\textbf{Rollout preferences.}
We evaluate each complete transition using the five quality axes in RAVEN~\cite{lu2026raven} and an additional terminal-image reward,
\begin{equation}
r_i^{\mathrm{end}}
=
-\operatorname{mean}
\operatorname{SmoothL1}\!\left(
S(I_{i,T})-S(I_T^\star)
\right),
\label{eq:method7}
\end{equation}
where \(I_{i,T}\) is the decoded frame at the actual global terminal index \(T\), and \(S\) is the resizing operator defined in Section~\ref{sec:trajectory}. The reward is evaluated once per complete rollout. Let \(q_{i,a}\) denote the normalized score on reward axis \(a\), covering the five quality axes and terminal-image agreement. The final rollout reward is
\begin{equation}
R_i
=
\frac{\sum_{a=1}^{6}w_a q_{i,a}}
{\sum_{a=1}^{6}|w_a|},
\label{eq:rollout_reward}
\end{equation}
where \(w_a\) weights axis \(a\). Group normalization and clipping of \(R_i\) yield the stopped advantage \(A_i\) used in policy learning.

\section{Experiments}
\label{sec:experiments}

\begin{table}[t]
\centering
\caption{\textbf{Main results on 168 general-purpose FL2VA cases.} Red shading from dark to light marks the best, second, and third distinct scores within each duration, with ties sharing a color. Stream. indicates the evaluated generation mode.}
\label{tab:all_results}
\setlength{\tabcolsep}{0.95pt}
\renewcommand{\arraystretch}{1.05}
\fontsize{7.4}{8.6}\selectfont
\begin{tabular*}{\textwidth}{@{\extracolsep{\fill}}lccccrrrrrrrrrr@{}}
\toprule
\multicolumn{6}{c}{} & \multicolumn{5}{c}{VABench} & \multicolumn{4}{c}{Video diagnostics} \\
\cmidrule(lr){7-11}\cmidrule(lr){12-15}
Model & Params. & Stream. & Audio & Steps & FPS & Lip-Sync $\uparrow$ & T-V Align $\uparrow$ & Expr. $\uparrow$ & Art. $\uparrow$ & VQA $\uparrow$ & Temp. $\uparrow$ & Color $\downarrow$ & Identity $\uparrow$ & \shortstack[r]{Final-frame\\PSNR (dB)$\uparrow$} \\
\midrule
\multicolumn{15}{l}{\textit{(a) 10s generation}} \\
H3 Base & 33B & No & Yes & 50 & 0.163 & \cellcolor{rankone}7.973 & \cellcolor{ranktwo}0.2134 & \cellcolor{rankone}4.018 & \cellcolor{ranktwo}3.351 & \cellcolor{rankone}0.995 & \cellcolor{rankthree}0.8051 & \cellcolor{rankthree}0.0109 & \cellcolor{ranktwo}0.9997 & \cellcolor{rankone}33.10 \\
H3 Turbo & 33B & No & Yes & 4 & \cellcolor{rankthree}0.950 & \cellcolor{rankthree}7.567 & 0.2090 & \cellcolor{rankthree}3.952 & 3.298 & 0.960 & 0.5791 & 0.0856 & 0.9994 & \cellcolor{ranktwo}32.58 \\
SkyReels & 14B & No & No & 30 & 0.269 & -- & 0.2101 & -- & -- & 0.929 & 0.7459 & 0.2008 & 0.9976 & 14.35 \\
Wan & 14B & No & No & 50 & 0.182 & -- & \cellcolor{rankthree}0.2109 & -- & -- & 0.933 & 0.5778 & 0.0313 & 0.9995 & 24.83 \\
OmniForcing & 33B & Yes & Yes & 3 & \cellcolor{ranktwo}1.333 & 0.964 & 0.2066 & 3.696 & 3.149 & 0.905 & 0.5834 & 0.0184 & 0.9986 & 29.98 \\
TaoMate & 33B & Yes & Yes & 3 & \cellcolor{rankone}1.795 & 7.235 & 0.2103 & 3.905 & \cellcolor{rankthree}3.339 & \cellcolor{rankthree}0.964 & \cellcolor{ranktwo}0.8800 & \cellcolor{ranktwo}0.0063 & \cellcolor{rankthree}0.9996 & 25.81 \\
\textbf{Ours} & 33B & Yes & Yes & 3 & \cellcolor{rankone}1.795 & \cellcolor{ranktwo}7.701 & \cellcolor{rankone}0.2176 & \cellcolor{ranktwo}3.988 & \cellcolor{rankone}3.393 & \cellcolor{ranktwo}0.977 & \cellcolor{rankone}0.8983 & \cellcolor{rankone}0.0058 & \cellcolor{rankone}0.9998 & \cellcolor{rankthree}31.14 \\
\midrule
\multicolumn{15}{l}{\textit{(b) 1min generation}} \\
H3 Base & 33B & No & Yes & 50 & 0.105 & 3.375 & \cellcolor{rankthree}0.2091 & \cellcolor{ranktwo}3.905 & \cellcolor{rankthree}3.423 & \cellcolor{ranktwo}0.968 & \cellcolor{rankthree}0.8084 & 0.0552 & 0.9987 & \cellcolor{rankone}32.70 \\
H3 Turbo & 33B & No & Yes & 4 & 0.866 & \cellcolor{rankthree}3.943 & 0.2021 & 3.756 & 3.387 & 0.895 & 0.7252 & 0.1512 & 0.9983 & \cellcolor{ranktwo}32.08 \\
SkyReels & 14B & No & No & 30 & 0.104 & -- & 0.2069 & -- & -- & 0.946 & 0.6174 & 0.1357 & 0.9920 & 13.62 \\
Wan & 14B & No & No & 50 & 0.143 & -- & 0.2041 & -- & -- & 0.865 & 0.6572 & 0.2163 & 0.9982 & 27.51 \\
OmniForcing & 33B & Yes & Yes & 3 & \cellcolor{rankthree}1.410 & 0.620 & 0.1987 & 3.435 & 3.173 & 0.926 & 0.4696 & \cellcolor{rankthree}0.0453 & \cellcolor{rankthree}0.9989 & 22.89 \\
TaoMate & 33B & Yes & Yes & 3 & \cellcolor{ranktwo}1.984 & \cellcolor{ranktwo}7.102 & \cellcolor{ranktwo}0.2165 & \cellcolor{rankthree}3.857 & \cellcolor{ranktwo}3.440 & \cellcolor{rankthree}0.958 & \cellcolor{ranktwo}0.8486 & \cellcolor{ranktwo}0.0078 & \cellcolor{ranktwo}0.9990 & 24.82 \\
\textbf{Ours} & 33B & Yes & Yes & 3 & \cellcolor{rankone}1.996 & \cellcolor{rankone}7.455 & \cellcolor{rankone}0.2198 & \cellcolor{rankone}3.929 & \cellcolor{rankone}3.464 & \cellcolor{rankone}0.981 & \cellcolor{rankone}0.9211 & \cellcolor{rankone}0.0076 & \cellcolor{rankone}0.9996 & \cellcolor{rankthree}28.92 \\
\bottomrule
\end{tabular*}
\end{table}

\subsection{Experimental setup}

\paragraph{Baselines and generation settings.}
We compare with MiniMax-H3 (H3 Base)~\cite{minimax2026h3modelcard}, H3 Turbo~\cite{lightx2v2026h3turbo}, SkyReels-V2-DF~\cite{skyreels2025repository}, and Wan FLF2V~\cite{wan2025repository}. SkyReels and Wan generate video only. OmniForcing~\cite{omniforcing2026} and TaoMate~\cite{gan2026taomate} are recent open-source streaming audiovisual distillation methods and our closest baselines. We retrain both for FL2VA. Our sampling and cache strategy follows TaoMate. H3 Base/Turbo and Wan have limited single-clip durations. We run SkyReels-V2-DF in its synchronous first/last-frame mode, denoising one clip at a time. Minute-long generation stitches six clips for H3 Base/Turbo and SkyReels, and twelve for Wan, using each preceding decoded final frame to join successive clips. Table~\ref{tab:all_results} reports generator sizes, inference steps, audio support, and throughput for a reference configuration of eight NVIDIA H20 GPUs.

\paragraph{Benchmark and metrics.}
Our general-purpose FL2VA benchmark spans 168 cases covering human activities, object interactions, and non-human subjects. At \textbf{10s} and \textbf{1min}, all methods generate \textbf{768p} videos from the same first and last images and audiovisual text conditions. VABench~\cite{hua2025vabench} measures Lip-Sync, text--video alignment (T-V Align, using ViCLIP), Expressiveness, Artistry, and VQA with shared annotations. We report temporal stability (Temp.), maximum scene-color deviation from the first frame (Color), and face-landmark consistency (Identity). At each duration, we average Identity over 136/168 cases with a visible human face (81.0\%). For audiovisual methods, we average Lip-Sync over 118/168 cases with visible lip movements during speech (70.2\%). We measure endpoint control by the PSNR between the last decoded frame and the requested final image. Higher is better except for Color. Dashes mark unavailable scores for video-only models, including Lip-Sync (which requires audio) and audiovisual Expressiveness and Artistry.

\begin{figure}[!htbp]
\centering
\includegraphics[width=\textwidth]{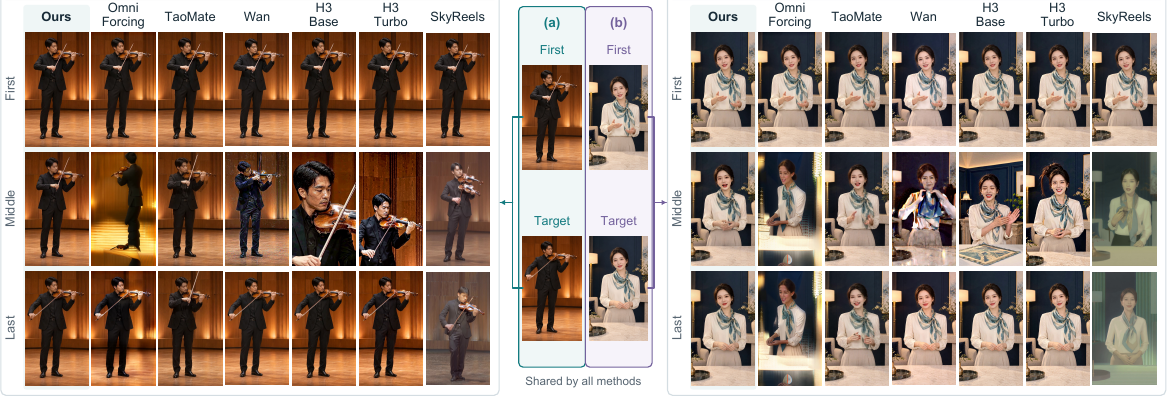}
\caption{\textbf{Minute-long visual comparisons.} The sampled frames show how our model follows each target pose while preserving the subject and framing.}
\label{fig:qualitative_comparison}
\end{figure}

\begin{figure}[!htbp]
\centering
\captionsetup{skip=3pt}
\includegraphics[width=.91\textwidth]{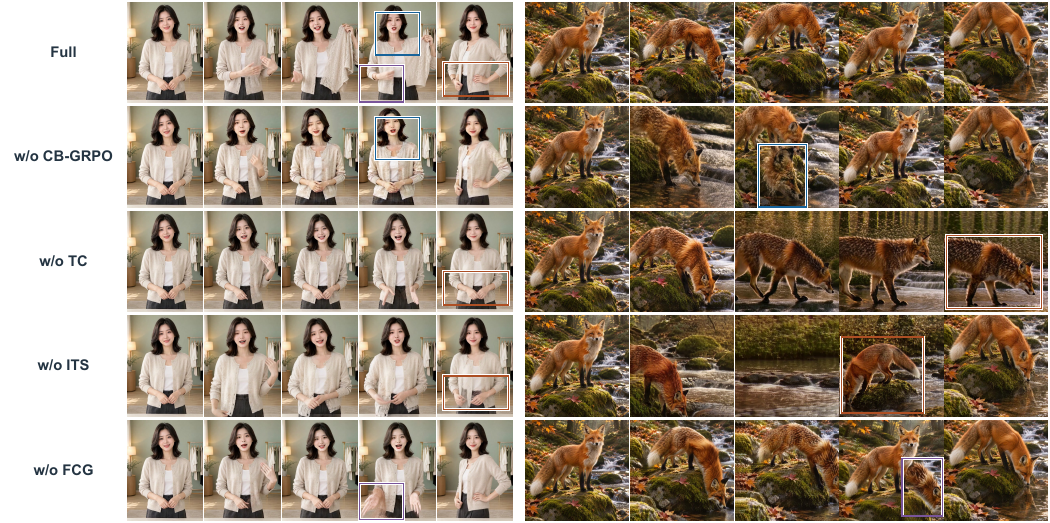}
\caption{\textbf{Qualitative ablation comparisons.} Colored boxes highlight intermediate-frame drift and final-frame deviations from the target. The full model maintains coherent intermediate content and accurate target alignment.}
\label{fig:ablation_sequences}
\end{figure}

\subsection{Quantitative results}

At 10s (Table~\ref{tab:all_results}), our model exceeds H3 Turbo in expressiveness (3.988) and surpasses both H3 variants in text--video alignment, artistry, temporal stability, color consistency, and facial landmark consistency. The advantage grows as generation extends beyond a single clip. At one minute, our method achieves the best or joint-best result on eight of the nine quality metrics, including all five VABench scores. Compared with H3 Base, it raises temporal stability from 0.8084 to 0.9211 and Lip-Sync from 3.375 to 7.455, while reducing color deviation from 0.0552 to 0.0076. Our three-step generator runs at 1.795 FPS for 10s and 1.996 FPS for one minute, about $11\times$ and $19\times$ the estimated H3 Base rates.

Our model outperforms both FL2VA-retrained streaming baselines on all nine quality metrics at both durations. With the same three-step sampler and cache strategy as TaoMate, it follows the final image more closely. Its final-frame PSNR reaches 31.14~dB at 10s and 28.92~dB at one minute, gains of 5.33 and 4.10~dB over TaoMate. The margin over OmniForcing widens from 1.16 to 6.03~dB as the stream grows longer.

\subsection{Visual comparison}

Figure~\ref{fig:qualitative_comparison} shows where stitched bidirectional clips lose continuity. In the violin sequence, the H3 variants move from a full-body view to a close shot midway. In the second sequence, Wan distorts the figure and alters the appearance of the scarf. A shared boundary image anchors the pose at a join, yet the next clip can send an ongoing gesture in a different direction. SkyReels introduces a visible color cast that changes the appearance of both subject and scene.

The streaming baselines also struggle to keep the intervening motion compatible with the target. In TaoMate's middle view, the hands are apart. They move closer together in the last view shown but remain higher than in the target image. OmniForcing drifts more noticeably midway through the violin sequence, where the violinist turns away from the requested view. Our model preserves the framing and each subject's appearance across the intermediate blocks while guiding the generation trajectory toward their respective target poses.

\subsection{Ablation studies}
\label{sec:ablations}

\paragraph{Ablation setup.}
Table~\ref{tab:ablation_modules} independently removes cross-block policy learning (CB-GRPO), temporal conditioning (TC), intermediate trajectory supervision (ITS), or full-chain denoising gradients (FCG) from the full model. The w/o TC variant receives the target image without its requested arrival time. Removing ITS eliminates intermediate spatial and temporal trajectory supervision by setting \(\lambda_s=\lambda_m=0\). Removing FCG randomly selects one denoising step for backpropagation and detaches preceding denoising states. Distillation variants share initialization, data, optimizer, batch size, and update budget. All except w/o CB-GRPO receive matched policy training. Evaluation uses the same 168 case--seed pairs per duration.

\begin{table}[t]
\centering
\caption{\textbf{Module ablations on the same 168 cases.} Each w/o removes one module from Full. Red shading from dark to light marks the best, second, and third distinct scores within each duration.}
\label{tab:ablation_modules}
\fontsize{8}{9.4}\selectfont
\setlength{\tabcolsep}{1.0pt}
\renewcommand{\arraystretch}{1.02}
\begin{tabular*}{\columnwidth}{@{\extracolsep{\fill}}lrrrrr@{}}
\toprule
 & \multicolumn{3}{c}{VABench} & \multicolumn{2}{c}{Video diagnostics} \\
\cmidrule(lr){2-4}\cmidrule(lr){5-6}
Variant & T-V Align$\uparrow$ & Lip-Sync$\uparrow$ & VQA$\uparrow$ & Temp.$\uparrow$ & \shortstack[r]{Final-frame\\PSNR (dB)$\uparrow$} \\
\midrule
\multicolumn{6}{l}{\textit{(a) 10s generation}} \\
Full & \cellcolor{rankone}0.2176 & \cellcolor{rankone}7.701 & \cellcolor{rankone}0.977 & \cellcolor{rankone}0.8983 & \cellcolor{rankone}31.14 \\
w/o CB-GRPO & 0.2112 & \cellcolor{ranktwo}7.633 & \cellcolor{rankthree}0.974 & 0.8634 & \cellcolor{ranktwo}31.04 \\
w/o TC & \cellcolor{ranktwo}0.2164 & \cellcolor{rankthree}7.534 & \cellcolor{ranktwo}0.976 & \cellcolor{ranktwo}0.8812 & 26.88 \\
w/o ITS & \cellcolor{rankthree}0.2158 & 7.343 & 0.971 & 0.8679 & 28.45 \\
w/o FCG & 0.2134 & 7.319 & 0.965 & \cellcolor{rankthree}0.8722 & \cellcolor{rankthree}30.85 \\
\midrule
\multicolumn{6}{l}{\textit{(b) 1min generation}} \\
Full & \cellcolor{rankone}0.2198 & \cellcolor{rankone}7.455 & \cellcolor{rankone}0.981 & \cellcolor{rankone}0.9211 & \cellcolor{rankone}28.92 \\
w/o CB-GRPO & 0.2168 & 7.320 & \cellcolor{rankthree}0.975 & 0.8740 & \cellcolor{ranktwo}28.73 \\
w/o TC & \cellcolor{ranktwo}0.2184 & \cellcolor{ranktwo}7.420 & \cellcolor{ranktwo}0.979 & \cellcolor{ranktwo}0.9106 & 25.89 \\
w/o ITS & 0.2177 & \cellcolor{rankthree}7.393 & 0.970 & \cellcolor{rankthree}0.9024 & 27.24 \\
w/o FCG & \cellcolor{rankthree}0.2180 & 7.238 & 0.967 & 0.8943 & \cellcolor{rankthree}28.61 \\
\bottomrule
\end{tabular*}
\end{table}

\paragraph{Temporal conditioning and trajectory supervision.}
Without TC's terminal-time control, uncorrected drift can prevent long videos from maintaining the target scene and subject, as illustrated by the fox in Figure~\ref{fig:ablation_sequences} (right). Endpoint PSNR is lowest at both durations, while minute-long stability falls from 0.9211 to 0.9106 despite similar T-V alignment (0.2184 vs.~0.2198), indicating weaker visual consistency. Removing ITS lowers PSNR by 2.69 and 1.68~dB and minute-long stability to 0.9024. Its fox sequence briefly loses definition before recovering the final drinking pose, illustrating the role of intermediate supervision in preserving coherent structure.

\paragraph{Full-chain denoising gradients.}
At one minute, full-chain denoising gradients raise temporal stability from 0.8943 to 0.9211 and Lip-Sync from 7.238 to 7.455 with the same forward sampler. In Figure~\ref{fig:ablation_sequences}, the single-step variant produces a translucent hand and overlapping fox bodies. An early denoising estimate establishes the spatial structure that later steps refine. Passing the loss through the full chain trains those early decisions before the block becomes part of the history seen by subsequent blocks.

\paragraph{Cross-block policy learning.}
Removing CB-GRPO produces the largest drop in temporal stability among the ablations. The gap widens from 0.0349 at 10s to 0.0471 at one minute, and minute-long VQA falls from 0.981 to 0.975. The full model also keeps lip movements better aligned with speech over a minute, scoring 7.455 against 7.320 without CB-GRPO. Without policy training, the face develops a second contour in Figure~\ref{fig:ablation_sequences}. A locally plausible pose can become an unreliable reference for later blocks. By propagating later policy losses through selected history paths, CB-GRPO trains early choices according to their effect on the continuation. FCG thus trains the denoising decisions that shape a block, while CB-GRPO also trains that block for the motion and appearance it supports later in the stream.

\section{Conclusion}

Taomate-F1 connects prescribed first and last images through streaming audiovisual generation. Temporal conditioning places the target image at its global terminal time, while paired teacher trajectories supervise intermediate structure and temporal change alongside endpoint reconstruction. Full-chain denoising gradients propagate this supervision through every retained sampling step within each block. CB-GRPO carries reward-weighted feedback from later blocks through selected history paths into earlier content generation. Experiments on our general-purpose FL2VA benchmark demonstrate improved minute-long audiovisual alignment and appearance consistency over the evaluated baselines. The resulting stream can be redirected through successive visual targets, supporting controllable long-form generation without prescribing the entire sequence \mbox{in advance.}

\FloatBarrier
\clearpage
\bibliographystyle{unsrtnat}
\bibliography{taomate_f1/references,taomate_f1/references_expansion,taomate_f1/references_experiments}

@inproceedings{zeng2023pixeldance,
  author = {Zeng, Yan and Wei, Guoqiang and Zheng, Jiani and Zou, Jiaxin and Wei, Yang and Zhang, Yuchen and Li, Hang},
  title = {{Make Pixels Dance: High-Dynamic Video Generation}},
  booktitle = {Proceedings of the IEEE/CVF Conference on Computer Vision and Pattern Recognition (CVPR)},
  pages = {8850--8860},
  year = {2024},
  doi = {10.1109/cvpr52733.2024.00845},
  url = {https://openaccess.thecvf.com/content/CVPR2024/html/Zeng_Make_Pixels_Dance_High-Dynamic_Video_Generation_CVPR_2024_paper.html}
}

@inproceedings{wang2024generativeinbetweening,
  author = {Wang, Xiaojuan and Zhou, Boyang and Curless, Brian and Kemelmacher-Shlizerman, Ira and Holynski, Aleksander and Seitz, Steven M.},
  title = {{Generative Inbetweening: Adapting Image-to-Video Models for Keyframe Interpolation}},
  booktitle = {International Conference on Learning Representations},
  year = {2025},
  url = {https://openreview.net/forum?id=ykD8a9gJvy}
}

@misc{minimax2026h3modelcard,
  author = {{MiniMax}},
  title = {{MiniMax H3}},
  year = {2026},
  url = {https://modelscope.cn/models/MiniMax/MiniMax-H3},
  howpublished = {Official model documentation}
}

@inproceedings{yin2024dmd,
  author = {Yin, Tianwei and Gharbi, Micha\"el and Zhang, Richard and Shechtman, Eli and Durand, Fr\'edo and Freeman, William T. and Park, Taesung},
  title = {{One-step Diffusion with Distribution Matching Distillation}},
  booktitle = {Proceedings of the IEEE/CVF Conference on Computer Vision and Pattern Recognition (CVPR)},
  pages = {6613--6623},
  year = {2024},
  doi = {10.1109/cvpr52733.2024.00632},
  url = {https://openaccess.thecvf.com/content/CVPR2024/html/Yin_One-step_Diffusion_with_Distribution_Matching_Distillation_CVPR_2024_paper.html}
}

@inproceedings{yin2024dmd2,
  author = {Yin, Tianwei and Gharbi, Micha\"{e}l and Park, Taesung and Zhang, Richard and Shechtman, Eli and Durand, Fr\'{e}do and Freeman, William T.},
  title = {{Improved Distribution Matching Distillation for Fast Image Synthesis}},
  booktitle = {Advances in Neural Information Processing Systems},
  volume = {37},
  pages = {47455--47487},
  publisher = {Curran Associates, Inc.},
  year = {2024},
  doi = {10.52202/079017-1505},
  url = {https://papers.nips.cc/paper_files/paper/2024/hash/54dcf25318f9de5a7a01f0a4125c541e-Abstract-Conference.html}
}

@inproceedings{huang2025selfforcing,
  author = {Huang, Xun and Li, Zhengqi and He, Guande and Zhou, Mingyuan and Shechtman, Eli},
  title = {{Self Forcing: Bridging the Train-Test Gap in Autoregressive Video Diffusion}},
  booktitle = {Advances in Neural Information Processing Systems},
  volume = {38},
  pages = {167283--167308},
  publisher = {Curran Associates, Inc.},
  year = {2025},
  doi = {10.52202/085713-5576},
  url = {https://proceedings.neurips.cc/paper_files/paper/2025/hash/f4823f831af67a3ef15e41a85434422a-Abstract-Conference.html}
}

@inproceedings{yang2025longlive,
  author = {Yang, Shuai and Huang, Wei and Chu, Ruihang and Xiao, Yicheng and Zhao, Yuyang and Wang, Xianbang and Li, Muyang and Xie, Enze and Chen, Yingcong and Lu, Yao and Han, Song and Chen, Yukang},
  title = {{LongLive: Real-time Interactive Long Video Generation}},
  booktitle = {International Conference on Learning Representations},
  year = {2026},
  url = {https://openreview.net/forum?id=nCAODkpsPJ}
}

@inproceedings{cui2025selfforcingplusplus,
  author = {Cui, Jiaxing and Wu, Jie and Li, Ming and Yang, Tao and Li, Xiaojie and Wang, Rui and Bai, Andrew and Ban, Yuanhao and Hsieh, Cho-Jui},
  title = {{Self-Forcing++: Towards Minute-Scale High-Quality Video Generation}},
  booktitle = {International Conference on Learning Representations},
  year = {2026},
  url = {https://proceedings.iclr.cc/paper_files/paper/2026/hash/8a30aba6514b56d02976f49797f6338a-Abstract-Conference.html}
}

@article{zhang2026connectedselfforcing,
  author = {Zhang, Dongbin and Zheng, Chaoda and Chen, Kangjie and Li, Xiangyu and Chen, Shijia and Deng, Jinhao and Zhang, Yuqi and Jiang, Guangfeng and Lin, Hongbin and {Choo Sin Wai} and Wang, Minqi and Wang, Puyi and Zhang, Jingye and Zhang, Yu and Liu, Xianming and Wang, Boyang},
  title = {{Connected Self Forcing: Beyond Local Learning in Video Autoregression}},
  journal = {arXiv preprint arXiv:2610.12156},
  year = {2026},
  doi = {10.48550/arXiv.2610.12156},
  url = {https://arxiv.org/abs/2610.12156v1}
}

@inproceedings{liu2025flowgrpo,
  author = {Liu, Jie and Liu, Gongye and Liang, Jiajun and Li, Yangguang and Liu, Jiaheng and Wang, Xintao and Wan, Pengfei and Zhang, Di and Ouyang, Wanli},
  title = {{Flow-GRPO: Training Flow Matching Models via Online RL}},
  booktitle = {Advances in Neural Information Processing Systems},
  volume = {38},
  pages = {40783--40818},
  publisher = {Curran Associates, Inc.},
  year = {2025},
  doi = {10.52202/085713-1362},
  url = {https://proceedings.neurips.cc/paper_files/paper/2025/hash/3a10c46572628d58cb44fb705f25cbbf-Abstract-Conference.html}
}

@inproceedings{zheng2025diffusionnft,
  author = {Zheng, Kaiwen and Chen, Huayu and Ye, Haotian and Wang, Haoxiang and Zhang, Qinsheng and Jiang, Kai and Su, Hang and Ermon, Stefano and Zhu, Jun and Liu, Ming-Yu},
  title = {{DiffusionNFT: Online Diffusion Reinforcement with Forward Process}},
  booktitle = {International Conference on Learning Representations},
  year = {2026},
  url = {https://openreview.net/forum?id=VJZ477R89F}
}

@article{lu2026raven,
  author = {Lu, Yanzuo and Zuo, Ronglai and Deng, Jiankang},
  title = {{RAVEN: Real-time Autoregressive Video Extrapolation with Consistency-model GRPO}},
  journal = {arXiv preprint arXiv:2605.15190},
  year = {2026},
  doi = {10.48550/arXiv.2605.15190},
  url = {https://arxiv.org/abs/2605.15190v1}
}

@inproceedings{schulman2015scg,
  author = {Schulman, John and Heess, Nicolas and Weber, Theophane and Abbeel, Pieter},
  title = {{Gradient Estimation Using Stochastic Computation Graphs}},
  booktitle = {Advances in Neural Information Processing Systems},
  volume = {28},
  publisher = {Curran Associates, Inc.},
  year = {2015},
  url = {https://papers.nips.cc/paper_files/paper/2015/hash/de03beffeed9da5f3639a621bcab5dd4-Abstract.html}
}

@inproceedings{ho2020ddpm,
  author = {Ho, Jonathan and Jain, Ajay and Abbeel, Pieter},
  title = {{Denoising Diffusion Probabilistic Models}},
  booktitle = {Advances in Neural Information Processing Systems},
  volume = {33},
  pages = {6840--6851},
  publisher = {Curran Associates, Inc.},
  year = {2020},
  url = {https://proceedings.neurips.cc/paper/2020/hash/4c5bcfec8584af0d967f1ab10179ca4b-Abstract.html}
}

@inproceedings{song2020ddim,
  author = {Song, Jiaming and Meng, Chenlin and Ermon, Stefano},
  title = {{Denoising Diffusion Implicit Models}},
  booktitle = {International Conference on Learning Representations},
  year = {2021},
  url = {https://openreview.net/forum?id=St1giarCHLP}
}

@inproceedings{rombach2021ldm,
  author = {Rombach, Robin and Blattmann, Andreas and Lorenz, Dominik and Esser, Patrick and Ommer, Bj\"orn},
  title = {{High-Resolution Image Synthesis With Latent Diffusion Models}},
  booktitle = {Proceedings of the IEEE/CVF Conference on Computer Vision and Pattern Recognition (CVPR)},
  pages = {10684--10695},
  year = {2022},
  doi = {10.1109/cvpr52688.2022.01042},
  url = {https://openaccess.thecvf.com/content/CVPR2022/html/Rombach_High-Resolution_Image_Synthesis_With_Latent_Diffusion_Models_CVPR_2022_paper.html}
}

@inproceedings{peebles2022dit,
  author = {Peebles, William and Xie, Saining},
  title = {{Scalable Diffusion Models with Transformers}},
  booktitle = {Proceedings of the IEEE/CVF International Conference on Computer Vision (ICCV)},
  pages = {4195--4205},
  year = {2023},
  doi = {10.1109/iccv51070.2023.00387},
  url = {https://openaccess.thecvf.com/content/ICCV2023/html/Peebles_Scalable_Diffusion_Models_with_Transformers_ICCV_2023_paper.html}
}

@inproceedings{ho2022videodiffusion,
  author = {Ho, Jonathan and Salimans, Tim and Gritsenko, Alexey and Chan, William and Norouzi, Mohammad and Fleet, David},
  title = {{Video Diffusion Models}},
  booktitle = {Advances in Neural Information Processing Systems},
  volume = {35},
  pages = {8633--8646},
  publisher = {Curran Associates, Inc.},
  year = {2022},
  doi = {10.52202/068431-0628},
  url = {https://proceedings.neurips.cc/paper_files/paper/2022/hash/39235c56aef13fb05a6adc95eb9d8d66-Abstract-Conference.html}
}

@inproceedings{blattmann2023videoldm,
  author = {Blattmann, Andreas and Rombach, Robin and Ling, Huan and Dockhorn, Tim and Kim, Seung Wook and Fidler, Sanja and Kreis, Karsten},
  title = {{Align Your Latents: High-Resolution Video Synthesis With Latent Diffusion Models}},
  booktitle = {Proceedings of the IEEE/CVF Conference on Computer Vision and Pattern Recognition (CVPR)},
  pages = {22563--22575},
  year = {2023},
  doi = {10.1109/cvpr52729.2023.02161},
  url = {https://openaccess.thecvf.com/content/CVPR2023/html/Blattmann_Align_Your_Latents_High-Resolution_Video_Synthesis_With_Latent_Diffusion_Models_CVPR_2023_paper.html}
}

@inproceedings{guo2023animatediff,
  author = {Guo, Yuwei and Yang, Ceyuan and Rao, Anyi and Liang, Zhengyang and Wang, Yaohui and Qiao, Yu and Agrawala, Maneesh and Lin, Dahua and Dai, Bo},
  title = {{AnimateDiff: Animate Your Personalized Text-to-Image Diffusion Models without Specific Tuning}},
  booktitle = {International Conference on Learning Representations},
  year = {2024},
  url = {https://openreview.net/forum?id=Fx2SbBgcte}
}

@inproceedings{yang2024cogvideox,
  author = {Yang, Zhuoyi and Teng, Jiayan and Zheng, Wendi and Ding, Ming and Huang, Shiyu and Xu, Jiazheng and Yang, Yuanming and Hong, Wenyi and Zhang, Xiaohan and Feng, Guanyu and Yin, Da and Zhang, Yuxuan and Wang, Weihan and Cheng, Yean and Xu, Bin and Gu, Xiaotao and Dong, Yuxiao and Tang, Jie},
  title = {{CogVideoX: Text-to-Video Diffusion Models with An Expert Transformer}},
  booktitle = {International Conference on Learning Representations},
  year = {2025},
  url = {https://openreview.net/forum?id=LQzN6TRFg9}
}

@article{zheng2025opensora2,
  author = {Zheng, Zangwei and Peng, Xiangyu and Lou, Yuxuan and Shen, Chenhui and Young, Tom and Guo, Xinying and Wang, Binluo and Xu, Hang and Liu, Hongxin and Jiang, Mingyan and Li, Wenjun and Wang, Yuhui and Ye, Anbang and Ren, Gang and Ma, Qianran and Liang, Wanying and Lian, Xiang and Wu, Xiwen and Zhong, Yuting and Li, Zhuangyan and Gong, Chaoyu and Lei, Guojun and Cheng, Leijun and Zhang, Limin and Li, Minghao and Zhang, Ruijie and Hu, Silan and Huang, Shijie and Wang, Xiaokang and Zhao, Yuanheng and Wang, Yuqi and Wei, Ziang and You, Yang},
  title = {{Open-Sora 2.0: Training a Commercial-Level Video Generation Model in \$200k}},
  journal = {arXiv preprint arXiv:2503.09642},
  year = {2025},
  doi = {10.48550/arXiv.2503.09642},
  url = {https://arxiv.org/abs/2503.09642v3}
}

@inproceedings{yin2024causvid,
  author = {Yin, Tianwei and Zhang, Qiang and Zhang, Richard and Freeman, William T. and Durand, Fredo and Shechtman, Eli and Huang, Xun},
  title = {{From Slow Bidirectional to Fast Autoregressive Video Diffusion Models}},
  booktitle = {Proceedings of the IEEE/CVF Conference on Computer Vision and Pattern Recognition (CVPR)},
  pages = {22963--22974},
  year = {2025},
  doi = {10.1109/cvpr52734.2025.02138},
  url = {https://openaccess.thecvf.com/content/CVPR2025/html/Yin_From_Slow_Bidirectional_to_Fast_Autoregressive_Video_Diffusion_Models_CVPR_2025_paper.html}
}

@inproceedings{chen2024diffusionforcing,
  author = {Chen, Boyuan and Mart\'{\i} Mons\'{o}, Diego and Du, Yilun and Simchowitz, Max and Tedrake, Russ and Sitzmann, Vincent},
  title = {{Diffusion Forcing: Next-token Prediction Meets Full-Sequence Diffusion}},
  booktitle = {Advances in Neural Information Processing Systems},
  volume = {37},
  pages = {24081--24125},
  publisher = {Curran Associates, Inc.},
  year = {2024},
  doi = {10.52202/079017-0759},
  url = {https://proceedings.neurips.cc/paper_files/paper/2024/hash/2aee1c4159e48407d68fe16ae8e6e49e-Abstract-Conference.html}
}

@inproceedings{singer2022makeavideo,
  author = {Singer, Uriel and Polyak, Adam and Hayes, Thomas and Yin, Xi and An, Jie and Zhang, Songyang and Hu, Qiyuan and Yang, Harry and Ashual, Oron and Gafni, Oran and Parikh, Devi and Gupta, Sonal and Taigman, Yaniv},
  title = {{Make-A-Video: Text-to-Video Generation without Text-Video Data}},
  booktitle = {International Conference on Learning Representations},
  year = {2023},
  url = {https://openreview.net/forum?id=nJfylDvgzlq}
}

@inproceedings{kondratyuk2023videopoet,
  author = {Kondratyuk, Dan and Yu, Lijun and Gu, Xiuye and Lezama, Jose and Huang, Jonathan and Schindler, Grant and Hornung, Rachel and Birodkar, Vighnesh and Yan, Jimmy and Chiu, Ming-Chang and Somandepalli, Krishna and Akbari, Hassan and Alon, Yair and Cheng, Yong and Dillon, Joshua V. and Gupta, Agrim and Hahn, Meera and Hauth, Anja and Hendon, David and Martinez, Alonso and Minnen, David and Sirotenko, Mikhail and Sohn, Kihyuk and Yang, Xuan and Adam, Hartwig and Yang, Ming-Hsuan and Essa, Irfan and Wang, Huisheng and Ross, David A and Seybold, Bryan and Jiang, Lu},
  title = {{V}ideo{P}oet: A Large Language Model for Zero-Shot Video Generation},
  booktitle = {Proceedings of the 41st International Conference on Machine Learning},
  volume = {235},
  pages = {25105--25124},
  publisher = {PMLR},
  year = {2024},
  url = {https://proceedings.mlr.press/v235/kondratyuk24a.html},
  series = {Proceedings of Machine Learning Research}
}

@inproceedings{salimans2022progressive,
  author = {Salimans, Tim and Ho, Jonathan},
  title = {{Progressive Distillation for Fast Sampling of Diffusion Models}},
  booktitle = {International Conference on Learning Representations},
  year = {2022},
  url = {https://openreview.net/forum?id=TIdIXIpzhoI}
}

@article{zhuang2026selfgradientforcing,
  author = {Zhuang, Junhao and Zhang, Shiyi and Bian, Yuxuan and Li, Yaowei and Luo, Yawen and Liu, Yijun and Jin, Weiyang and Zhang, Songchun and He, Xianglong and Zhang, Xuying and Li, Haoran and Huang, Haoyang and Xue, Zeyue and Duan, Nan},
  title = {{Self Gradient Forcing: Native Long Video Extrapolation}},
  journal = {arXiv preprint arXiv:2607.20368},
  year = {2026},
  doi = {10.48550/arXiv.2607.20368},
  url = {https://arxiv.org/abs/2607.20368}
}

@article{zhang2026vidus2,
  author = {Zhang, Jintao and Jiang, Kai and Chen, Jintao and Wang, Xu and Liu, Deyuan and Li, Jungang and Chen, Dechuang and Lin, Ming and Zhou, Jingjiang and Jin, Haopeng and Jia, Qi and Wang, Xiaohang and Wang, Yaole and Zhang, Zhanqiang and Li, Ran and Huang, Zhengkun and Xiong, Shuyue and Wang, Yuji and Dai, Zikun and He, Hui and Luo, Yang and Ning, Mang and Feng, Weiqi and Ye, Chengyang and Lin, Xinyue and Zhao, Min and Zhu, Hongzhou and Tan, Hengkai and Wang, Zeyuan and Xiang, Chendong and Zheng, Kaiwen and Deng, Zhijie and Bao, Fan and Chen, Jianfei and Zhu, Jun},
  title = {{Vidu S2: Real-Time Interactive, Editable, and Spatial Video Generation}},
  journal = {arXiv preprint arXiv:2609.11638},
  year = {2026},
  doi = {10.48550/arXiv.2609.11638},
  url = {https://arxiv.org/abs/2609.11638}
}

@article{gan2026taomate,
  author = {Gan, Qijun and Zhang, Chenwei and Jin, Meiguang and Ma, Junfeng and Shen, Qiu},
  title = {{TaoMate: Anchor-Guided Memory Bridging Evolving and Reference States for Real-Time Audio-Video Digital Human Generation}},
  journal = {arXiv preprint arXiv:2607.24359},
  year = {2026},
  doi = {10.48550/arXiv.2607.24359},
  url = {https://arxiv.org/abs/2607.24359}
}

@article{xing2024tooncrafter,
  author = {Xing, Jinbo and Liu, Hanyuan and Xia, Menghan and Zhang, Yong and Wang, Xintao and Shan, Ying and Wong, Tien-Tsin},
  title = {{ToonCrafter: Generative Cartoon Interpolation}},
  journal = {ACM Transactions on Graphics},
  volume = {43},
  number = {6},
  pages = {1--11},
  year = {2024},
  doi = {10.1145/3687761},
  url = {https://doi.org/10.1145/3687761}
}

@inproceedings{xing2023dynamicrafter,
  author = {Xing, Jinbo and Xia, Menghan and Zhang, Yong and Chen, Haoxin and Yu, Wangbo and Liu, Hanyuan and Liu, Gongye and Wang, Xintao and Shan, Ying and Wong, Tien-Tsin},
  title = {{DynamiCrafter: Animating Open-Domain Images with Video Diffusion Priors}},
  booktitle = {Computer Vision -- ECCV 2024},
  pages = {399--417},
  year = {2024},
  doi = {10.1007/978-3-031-72952-2_23},
  url = {https://doi.org/10.1007/978-3-031-72952-2_23}
}

@inproceedings{chen2023seine,
  author = {Chen, Xinyuan and Wang, Yaohui and Zhang, Lingjun and Zhuang, Shaobin and Ma, Xin and Yu, Jiashuo and Wang, Yali and Lin, Dahua and Qiao, Yu and Liu, Ziwei},
  title = {{SEINE: Short-to-Long Video Diffusion Model for Generative Transition and Prediction}},
  booktitle = {International Conference on Learning Representations},
  year = {2024},
  url = {https://openreview.net/forum?id=FNq3nIvP4F}
}

@inproceedings{guo2023sparsectrl,
  author = {Guo, Yuwei and Yang, Ceyuan and Rao, Anyi and Agrawala, Maneesh and Lin, Dahua and Dai, Bo},
  title = {{SparseCtrl: Adding Sparse Controls to Text-to-Video Diffusion Models}},
  booktitle = {Computer Vision -- ECCV 2024},
  pages = {330--348},
  year = {2024},
  doi = {10.1007/978-3-031-72946-1_19},
  url = {https://doi.org/10.1007/978-3-031-72946-1_19}
}

@article{polyak2024moviegen,
  author = {Polyak, Adam and Zohar, Amit and Brown, Andrew and Tjandra, Andros and Sinha, Animesh and Lee, Ann and Vyas, Apoorv and Shi, Bowen and Ma, Chih-Yao and Chuang, Ching-Yao and Yan, David and Choudhary, Dhruv and Wang, Dingkang and Sethi, Geet and Pang, Guan and Ma, Haoyu and Misra, Ishan and Hou, Ji and Wang, Jialiang and Jagadeesh, Kiran and Li, Kunpeng and Zhang, Luxin and Singh, Mannat and Williamson, Mary and Le, Matt and Yu, Matthew and Singh, Mitesh Kumar and Zhang, Peizhao and Vajda, Peter and Duval, Quentin and Girdhar, Rohit and Sumbaly, Roshan and Rambhatla, Sai Saketh and Tsai, Sam and Azadi, Samaneh and Datta, Samyak and Chen, Sanyuan and Bell, Sean and Ramaswamy, Sharadh and Sheynin, Shelly and Bhattacharya, Siddharth and Motwani, Simran and Xu, Tao and Li, Tianhe and Hou, Tingbo and Hsu, Wei-Ning and Yin, Xi and Dai, Xiaoliang and Taigman, Yaniv and Luo, Yaqiao and Liu, Yen-Cheng and Wu, Yi-Chiao and Zhao, Yue and Kirstain, Yuval and He, Zecheng and He, Zijian and Pumarola, Albert and Thabet, Ali and Sanakoyeu, Artsiom and Mallya, Arun and Guo, Baishan and Araya, Boris and Kerr, Breena and Wood, Carleigh and Liu, Ce and Peng, Cen and Vengertsev, Dimitry and Schonfeld, Edgar and Blanchard, Elliot and Juefei-Xu, Felix and Nord, Fraylie and Liang, Jeff and Hoffman, John and Kohler, Jonas and Fire, Kaolin and Sivakumar, Karthik and Chen, Lawrence and Yu, Licheng and Gao, Luya and Georgopoulos, Markos and Moritz, Rashel and Sampson, Sara K. and Li, Shikai and Parmeggiani, Simone and Fine, Steve and Fowler, Tara and Petrovic, Vladan and Du, Yuming},
  title = {{Movie Gen: A Cast of Media Foundation Models}},
  journal = {arXiv preprint arXiv:2410.13720},
  year = {2024},
  doi = {10.48550/arXiv.2410.13720},
  url = {https://arxiv.org/abs/2410.13720}
}

@inproceedings{ruan2022mmdiffusion,
  author = {Ruan, Ludan and Ma, Yiyang and Yang, Huan and He, Huiguo and Liu, Bei and Fu, Jianlong and Yuan, Nicholas Jing and Jin, Qin and Guo, Baining},
  title = {{MM-Diffusion: Learning Multi-Modal Diffusion Models for Joint Audio and Video Generation}},
  booktitle = {Proceedings of the IEEE/CVF Conference on Computer Vision and Pattern Recognition (CVPR)},
  pages = {10219--10228},
  year = {2023},
  doi = {10.1109/cvpr52729.2023.00985},
  url = {https://openaccess.thecvf.com/content/CVPR2023/html/Ruan_MM-Diffusion_Learning_Multi-Modal_Diffusion_Models_for_Joint_Audio_and_Video_CVPR_2023_paper.html}
}

@inproceedings{wang2024avdit,
  author = {Wang, Kai and Deng, Shijian and Shi, Jing and Hatzinakos, Dimitrios and Tian, Yapeng},
  title = {{AV-DiT: Taming Image Diffusion Transformers for Efficient Joint Audio and Video Generation}},
  booktitle = {Proceedings of the 33rd ACM International Conference on Multimedia},
  pages = {10486--10495},
  year = {2025},
  doi = {10.1145/3746027.3755713},
  url = {https://doi.org/10.1145/3746027.3755713}
}

@inproceedings{tian2024emo,
  author = {Tian, Linrui and Wang, Qi and Zhang, Bang and Bo, Liefeng},
  title = {{EMO: Emote Portrait Alive Generating Expressive Portrait Videos with Audio2Video Diffusion Model Under Weak Conditions}},
  booktitle = {Computer Vision -- ECCV 2024},
  pages = {244--260},
  year = {2024},
  doi = {10.1007/978-3-031-73010-8_15},
  url = {https://doi.org/10.1007/978-3-031-73010-8_15}
}

@inproceedings{xu2024vasa,
  author = {Xu, Sicheng and Chen, Guojun and Guo, Yu-Xiao and Yang, Jiaolong and Li, Chong and Zang, Zhenyu and Zhang, Yizhong and Tong, Xin and Guo, Baining},
  title = {{VASA-1: Lifelike Audio-Driven Talking Faces Generated in Real Time}},
  booktitle = {Advances in Neural Information Processing Systems},
  volume = {37},
  pages = {660--684},
  year = {2024},
  doi = {10.52202/079017-0021},
  url = {https://proceedings.neurips.cc/paper_files/paper/2024/hash/014fe398da515cd552fa6e1f33e0565e-Abstract-Conference.html}
}

@article{xu2024hallo,
  author = {Xu, Mingwang and Li, Hui and Su, Qingkun and Shang, Hanlin and Zhang, Liwei and Liu, Ce and Wang, Jingdong and Yao, Yao and Zhu, Siyu},
  title = {{Hallo: Hierarchical Audio-Driven Visual Synthesis for Portrait Image Animation}},
  journal = {arXiv preprint arXiv:2406.08801},
  year = {2024},
  doi = {10.48550/arXiv.2406.08801},
  url = {https://arxiv.org/abs/2406.08801}
}

@inproceedings{black2023ddpo,
  author = {Black, Kevin and Janner, Michael and Du, Yilun and Kostrikov, Ilya and Levine, Sergey},
  title = {{Training Diffusion Models with Reinforcement Learning}},
  booktitle = {International Conference on Learning Representations},
  year = {2024},
  url = {https://openreview.net/forum?id=YCWjhGrJFD}
}

@inproceedings{fan2023dpok,
  author = {Fan, Ying and Watkins, Olivia and Du, Yuqing and Liu, Hao and Ryu, Moonkyung and Boutilier, Craig and Abbeel, Pieter and Ghavamzadeh, Mohammad and Lee, Kangwook and Lee, Kimin},
  title = {{DPOK: Reinforcement Learning for Fine-tuning Text-to-Image Diffusion Models}},
  booktitle = {Advances in Neural Information Processing Systems},
  volume = {36},
  pages = {79858--79885},
  year = {2023},
  doi = {10.52202/075280-3497},
  url = {https://proceedings.neurips.cc/paper_files/paper/2023/hash/fc65fab891d83433bd3c8d966edde311-Abstract-Conference.html}
}

@article{prabhudesai2023alignprop,
  author = {Prabhudesai, Mihir and Goyal, Anirudh and Pathak, Deepak and Fragkiadaki, Katerina},
  title = {{Aligning Text-to-Image Diffusion Models with Reward Backpropagation}},
  journal = {arXiv preprint arXiv:2310.03739},
  year = {2023},
  doi = {10.48550/arXiv.2310.03739},
  url = {https://arxiv.org/abs/2310.03739}
}

@inproceedings{wallace2023diffusiondpo,
  author = {Wallace, Bram and Dang, Meihua and Rafailov, Rafael and Zhou, Linqi and Lou, Aaron and Purushwalkam, Senthil and Ermon, Stefano and Xiong, Caiming and Joty, Shafiq and Naik, Nikhil},
  title = {{Diffusion Model Alignment Using Direct Preference Optimization}},
  booktitle = {Proceedings of the IEEE/CVF Conference on Computer Vision and Pattern Recognition (CVPR)},
  pages = {8228--8238},
  year = {2024},
  doi = {10.1109/cvpr52733.2024.00786},
  url = {https://openaccess.thecvf.com/content/CVPR2024/html/Wallace_Diffusion_Model_Alignment_Using_Direct_Preference_Optimization_CVPR_2024_paper.html}
}

@article{shao2024deepseekmath,
  author = {Shao, Zhihong and Wang, Peiyi and Zhu, Qihao and Xu, Runxin and Song, Junxiao and Bi, Xiao and Zhang, Haowei and Zhang, Mingchuan and Li, Y. K. and Wu, Y. and Guo, Daya},
  title = {{DeepSeekMath: Pushing the Limits of Mathematical Reasoning in Open Language Models}},
  journal = {arXiv preprint arXiv:2402.03300},
  year = {2024},
  doi = {10.48550/arXiv.2402.03300},
  url = {https://arxiv.org/abs/2402.03300}
}

@article{schulman2017ppo,
  author = {Schulman, John and Wolski, Filip and Dhariwal, Prafulla and Radford, Alec and Klimov, Oleg},
  title = {{Proximal Policy Optimization Algorithms}},
  journal = {arXiv preprint arXiv:1707.06347},
  year = {2017},
  doi = {10.48550/arXiv.1707.06347},
  url = {https://arxiv.org/abs/1707.06347}
}

@inproceedings{song2023consistency,
  author = {Song, Yang and Dhariwal, Prafulla and Chen, Mark and Sutskever, Ilya},
  title = {{Consistency Models}},
  booktitle = {Proceedings of the 40th International Conference on Machine Learning},
  volume = {202},
  pages = {32211--32252},
  year = {2023},
  url = {https://proceedings.mlr.press/v202/song23a.html},
  series = {Proceedings of Machine Learning Research}
}

@article{luo2023lcm,
  author = {Luo, Simian and Tan, Yiqin and Huang, Longbo and Li, Jian and Zhao, Hang},
  title = {{Latent Consistency Models: Synthesizing High-Resolution Images with Few-Step Inference}},
  journal = {arXiv preprint arXiv:2310.04378},
  year = {2023},
  doi = {10.48550/arXiv.2310.04378},
  url = {https://arxiv.org/abs/2310.04378}
}

@inproceedings{kim2022diffusionclip,
  author = {Kim, Gwanghyun and Kwon, Taesung and Ye, Jong Chul},
  title = {{DiffusionCLIP: Text-Guided Diffusion Models for Robust Image Manipulation}},
  booktitle = {Proceedings of the IEEE/CVF Conference on Computer Vision and Pattern Recognition},
  year = {2022},
  url = {https://arxiv.org/abs/2110.02711v6}
}

@inproceedings{clark2024draft,
  author = {Clark, Kevin and Vicol, Paul and Swersky, Kevin and Fleet, David J.},
  title = {{Directly Fine-Tuning Diffusion Models on Differentiable Rewards}},
  booktitle = {International Conference on Learning Representations},
  year = {2024},
  url = {https://arxiv.org/abs/2309.17400v2}
}

@article{su2024roformer,
  author = {Su, Jianlin and Ahmed, Murtadha and Lu, Yu and Pan, Shengfeng and Bo, Wen and Liu, Yunfeng},
  title = {{RoFormer}: Enhanced Transformer with Rotary Position Embedding},
  journal = {Neurocomputing},
  volume = {568},
  pages = {127063},
  year = {2024},
  doi = {10.1016/j.neucom.2023.127063}
}

@article{omniforcing2026,
  author = {Su, Yaofeng and Li, Yuming and Xue, Zeyue and Huang, Jie and Fu, Siming and Li, Haoran and Li, Ying and Qian, Zezhong and Huang, Haoyang and Duan, Nan},
  title = {{OmniForcing}: Unleashing Real-time Joint Audio-Visual Generation},
  journal = {arXiv preprint arXiv:2603.11647},
  year = {2026},
  doi = {10.48550/arXiv.2603.11647},
  url = {https://arxiv.org/abs/2603.11647}
}

@misc{skyreels2025repository,
  author = {{SkyReels Team}},
  title = {{SkyReels V2}: Infinite-Length Film Generative Model},
  year = {2025},
  url = {https://github.com/SkyworkAI/SkyReels-V2},
  howpublished = {Official model and inference repository}
}

@misc{wan2025repository,
  author = {{Wan Team}},
  title = {{Wan2.1}: Open and Advanced Large-Scale Video Generative Models},
  year = {2025},
  url = {https://github.com/Wan-Video/Wan2.1},
  howpublished = {Official model and inference repository}
}

@inproceedings{hua2025vabench,
  author = {Hua, Daili and Wang, Xizhi and Zeng, Bohan and Huang, Xinyi and Liang, Hao and Niu, Junbo and Chen, Xinlong and Xu, Quanqing and Zhang, Wentao},
  title = {{VABench}: A Comprehensive Benchmark for Audio-Video Generation},
  booktitle = {Proceedings of the IEEE/CVF Conference on Computer Vision and Pattern Recognition (CVPR)},
  pages = {23345--23355},
  year = {2026},
  url = {https://openaccess.thecvf.com/content/CVPR2026/html/Hua_VABench_A_Comprehensive_Benchmark_for_Audio-Video_Generation_CVPR_2026_paper.html}
}

@article{wan2025technical,
  author = {{Team Wan} and others},
  title = {Wan: Open and Advanced Large-Scale Video Generative Models},
  journal = {arXiv preprint arXiv:2503.20314},
  year = {2025},
  doi = {10.48550/arXiv.2503.20314},
  url = {https://arxiv.org/abs/2503.20314}
}

@article{chen2025skyreelsv2,
  author = {Chen, Guibin and Lin, Dixuan and Yang, Jiangping and Lin, Chunze and Zhu, Junchen and Fan, Mingyuan and Zhang, Hao and Chen, Sheng and Chen, Zheng and Ma, Chengcheng and Xiong, Weiming and Wang, Wei and Pang, Nuo and Kang, Kang and Xu, Zhiheng and Jin, Yuzhe and Liang, Yupeng and Song, Yubing and Zhao, Peng and Xu, Boyuan and Qiu, Di and Li, Debang and Fei, Zhengcong and Li, Yang and Zhou, Yahui},
  title = {{SkyReels-V2}: Infinite-length Film Generative Model},
  journal = {arXiv preprint arXiv:2504.13074},
  year = {2025},
  doi = {10.48550/arXiv.2504.13074},
  url = {https://arxiv.org/abs/2504.13074}
}

@misc{lightx2v2026h3turbo,
  author = {{LightX2V Team}},
  title = {{MiniMax-H3 Turbo}},
  year = {2026},
  howpublished = {Official model release},
  url = {https://huggingface.co/lightx2v/Minimax-h3-Turbo}
}
\end{document}